\documentclass[10pt,twocolumn,letterpaper]{article}

\usepackage[pagenumbers]{cvpr} 

\usepackage{graphicx}
\usepackage{amsmath}
\usepackage{amssymb}
\usepackage{booktabs}
\usepackage{listings}
\usepackage{algorithm}
\usepackage{algpseudocode}
\usepackage{xcolor}

\usepackage[pagebackref,breaklinks,colorlinks]{hyperref}

\usepackage[capitalize]{cleveref}
\crefname{section}{Sec.}{Secs.}
\Crefname{section}{Section}{Sections}
\Crefname{table}{Table}{Tables}
\crefname{table}{Tab.}{Tabs.}

\def\cvprPaperID{****} 
\def\confName{CVPR}
\def\confYear{2023}

\begin{document}

\title{ Masked Contrastive Representation Learning}

\author{Yuchong Yao\\
The University of Melbourne\\
Parkville VIC 3010\\
{\tt\small yuchongy1@student.unimelb.edu.au}
\and
Nandakishor Desai\\
The University of Melbourne\\
Parkville VIC 3010\\
{\tt\small nandakishor.desai@unimelb.edu.au}
\and
Marimuthu Palaniswami\\
The University of Melbourne\\
Parkville VIC 3010\\
{\tt\small palani@unimelb.edu.au}
}
\maketitle

\begin{abstract}
Masked image modelling (e.g., Masked AutoEncoder) and contrastive learning (e.g., Momentum Contrast) have shown impressive performance on unsupervised visual representation learning. This work presents Masked Contrastive Representation Learning (MACRL) for self-supervised visual pre-training. In particular, MACRL leverages the effectiveness of both masked image modelling and contrastive learning. We adopt an asymmetric setting for the siamese network (i.e., encoder-decoder structure in both branches), where one branch with higher mask ratio and stronger data augmentation, while the other adopts weaker data corruptions. We optimize a contrastive learning objective based on the learned features from the encoder in both branches. Furthermore, we minimize the $L_1$ reconstruction loss according to the decoders' outputs. In our experiments, MACRL presents superior results on various vision benchmarks, including CIFAR-10, CIFAR-100, Tiny-ImageNet, and two other ImageNet subsets. Our framework provides unified insights on self-supervised visual pre-training and future research.

\end{abstract}

\section{Introduction}
\label{sec:intro}

Deep learning \cite{lecun2015deep} has demonstrated exceptional performance in the past years, showing dominant results in various tasks and applications. Modern architectures \cite{he2016deep, dosovitskiy2020image} can extract meaningful representations from millions of data entries, which are commonly labelled. With the explosion of available data resources, larger and deeper models could be established to obtain better generalizability and serve as foundation models for the downstream tasks \cite{devlin2018bert, bao2021beit}. However, as there are only limited annotations in the data, this encourages the models to learn in an unsupervised (i.e., self-supervised) fashion.

In Computer Vision, contrastive learning was the de-facto and dominant self-supervised learning paradigm for large-scale pre-training \cite{wu2018unsupervised}. Momentum Contrast (MoCo) \cite{he2020momentum} and SimCLR \cite{chen2020simple} are two signature contrastive learning methods, which adopt siamse network structure to maximize the agreement of learned representations between similar samples. They rely on strong data augmentations and scale well with the size of the data.

On the other side, in Natural Language Processing, this self-supervised learning problem is usually addressed by masked language modelling \cite{radford2018improving, devlin2018bert} in either autoregressive or autoencoding style. The core idea is to minimize the reconstruction loss of corrupted masked sentences. This simple but effective approach enables the training of large-scale language models that generalize well on various tasks. More recently, masked modelling has been generalized to Computer Vision (i.e., masked image modelling), where the models are expected to reconstruct masked image patches in either autoregressive or autoencoding manner \cite{chen2020generative, bao2021beit}. Masked AutoEncoder (MAE) \cite{he2022masked} is one of the most influential works in masked image modelling for its simple design and excellent efficiency. The experiment results suggest that masked image modelling surpasses the performance of contrastive learning and has become the new state-of-the-art approach for self-supervised visual pre-training.

However, masked image modelling still serves limitations. Without a pre-trained tokenizer as in iBOT \cite{zhou2021ibot} or knowledge distillation from pre-trained checkpoints \cite{liu2022exploring}, there is a gap in the linear probe accuracy between masked image modelling and contrastive learning, where the latter has better accuracy. Additionally, masked image modelling optimizes the pixel-level reconstruction objective, which lacks semantic information regarding the learned features. On the contrary, contrastive learning emphasizes the higher feature level similarity (or dissimilarity), leading to more semantic meaningful representations. Therefore, we would like to ask one question: \textbf{\textit{Is it possible to combine the merits of both masked image modelling and contrastive learning? }}

Driven by this motivation, we present Masked Contrastive Representation Learning (MACRL) for self-supervised visual pre-training. Inspired by \cite{wang2022importance}, MACRL adopts an asymmetric siamese network structure. One branch applies stronger data augmentations and a higher mask ratio for the images (defined as the main branch), while the other uses weaker data augmentations and no masks. Each branch consists of an asymmetric encoder-decoder network, resembling the design in MAE. The encoder is a vision transformer, and the decoder is a much shallow (e.g., 2-layer) network with a linear layer and an attention layer. Following MoCo, the main branch is updated using gradient propagation, and the other branch is updated with momentum. There is an additional projection head and momentum projection head for each branch, respectively, which also follow the MoCo updating convention. Overall, MACRL optimizes a reconstruction loss and a symmetric constructive objective. The former is based on the decoder output, and the latter follows the encoder's and momentum encoder's output after the (momentum) projection head.

Our MACRL obtains meaningful representations from unlabelled data in both pixel-level details and high-level semantics. We achieved superior fine-tune and linear probe accuracy across multiple vision benchmarks. Moreover, MACRL presents better efficiency in representation learning, which captures the semantic information more easily and within less training epochs. Furthermore, it shows better interpretability over two existing state-of-the-art methods. The observations encourage the community explore the connection and complementary between masked modelling and contrastive learning.

\begin{figure*}
  \centering
  \begin{subfigure}{0.58\linewidth}
    \centering
    \includegraphics[width=0.8\textwidth, height=4.5cm]{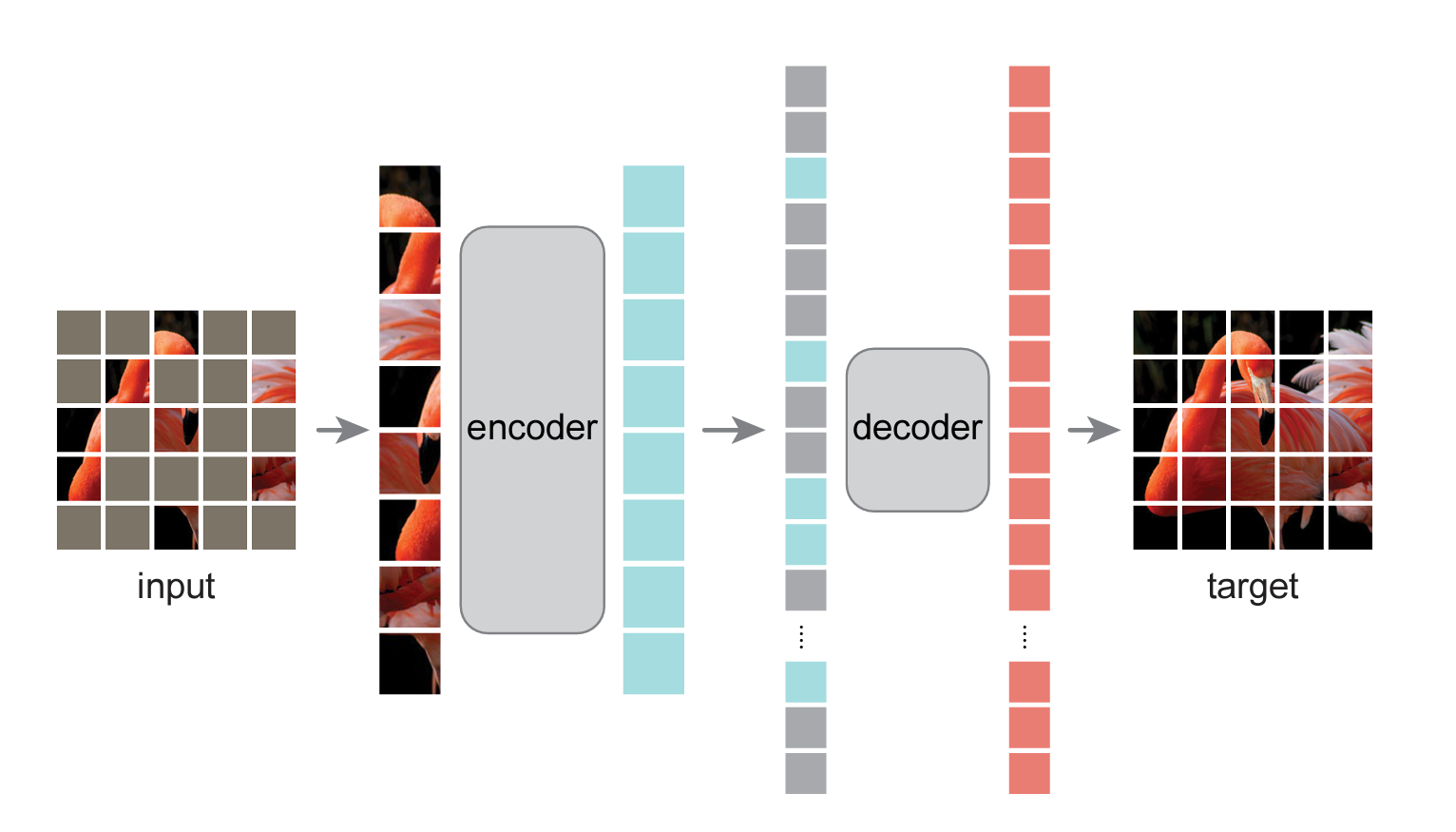}
    \caption{Example of Masked Image Modelling \cite{he2022masked}}
    
    \label{fig:mae}
  \end{subfigure}
  \hfill
  \begin{subfigure}{0.38\linewidth}
  \centering
    \includegraphics[width=0.8\textwidth, height=4cm]{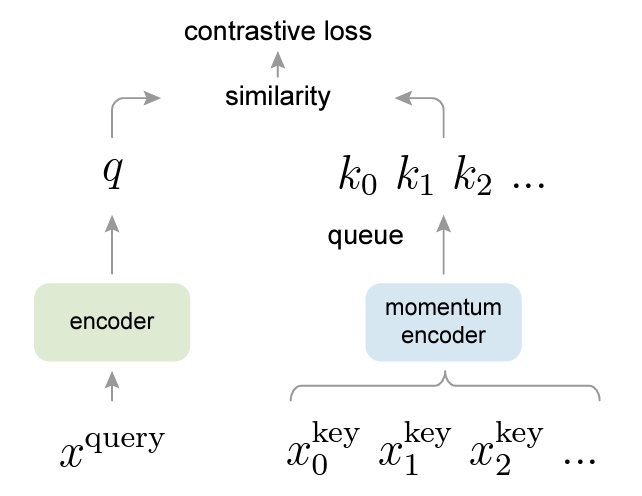}
    \caption{Example of Contrastive Learning \cite{he2020momentum}}
    \label{fig:moco}
  \end{subfigure}
  \caption{\textbf{Two Mainstream Self-Supervised Learning Approaches}. \textbf{Figure \ref{fig:mae}} shows the Masked Autoencoder (MAE), which minimizes the pixel reconstruction loss from randomly masked images. \textbf{Figure \ref{fig:moco}} illustrates the Momentum Contrast (MoCo), relying on the power of strong data augmentation, momentum encoder, memory bank etc. It learns representations by maximizing the agreement between similar representations.}
  \label{fig:ssl}
\end{figure*}

\section{Related Work}
\label{sec:related}

Masked Image Modelling and Contrastive Learning are two mainstream Self-Supervised Learning approaches (see \textbf{Figure \ref{fig:ssl}}), showing exceptional performance in recent years.

\textbf{Masked Image Modelling.} The idea of masked image modelling could be traced back to the early work in \cite{pathak2016context}, which is also pretext-based learning. In iGPT \cite{chen2020generative} presented GPT style pre-training on images, with an autoregressive prediction and an autoencoding denoising objective. BEiT \cite{bao2021beit} followed the BERT style pre-training with minor modifications. It first learned a tokenizer (dVAE) trained according to the principle in \cite{ramesh2021zero} (i.e., autoencoding style reconstruction). The tokenizer transformed the image into visual tokens. During pre-training, the task is to predict the visual token of the original image based on the encoded representations from the masked image. Such training settings helped BEiT outperform previous contrastive learning state-of-the-art in finetuning (e.g., MoCo v3 \cite{chen2021empirical}, DINO \cite{caron2021emerging}). MAE \cite{he2022masked} (see \textbf{Figure \ref{fig:mae}}) is another very influential work in masked modelling. The method is very simple that it predicts the randomly masked pixels but achieved very powerful results (e.g., surpassed BEiT and supervised approaches by a large margin). The authors developed an asymmetric encoder-decoder architecture, where the encoder only processes the unmasked patches, and the decoder operates on both encoded unmasked patches and masked patches. The objective is the mean square loss between the normalized reconstructed image and the original image. The work showed that random masking is very effective and could even work when the masking ratio is as high as 75\% (the original masking ratio in BERT is 15\%). It also suggested that images contain a lot of redundant information. With the lightweight decoder and high masking ratio, MAE is computationally efficient and scales well. SimMIM \cite{xie2022simmim} is a concurrent work with MAE and achieved similar results, with additional support for hierarchical vision transformer (e.g., Swin \cite{liu2021swin}). iBOT \cite{zhou2021ibot} presented an online tokenizer trained by forcing the similarity between cross-view images. The tokenizer was jointly optimized with masked image modelling with momentum update and self-distillation. dBOT \cite{liu2022exploring} also utilized knowledge distillation with masked image modelling by learning from data-richer teachers (e.g., CLIP \cite{radford2021learning}). \cite{dong2021peco, shi2022adversarial, assran2022masked, kakogeorgiou2022hide, li2021mst, li2022uniform} attempted to improve masked image modelling by either working on intermediate tokens/layers or introducing more advanced masking strategy. In \cite{xie2022revealing}, the authors studied why masked modelling achieved better results than supervised pre-training. They found that masked image modelling introduced locality inductive bias to all layers, which is critical for ViTs. Moreover, masked modelling gives diverse attention heads in all layers.

\textbf{Contrastive Learning.} The idea can be traced back to instance discrimination \cite{wu2018unsupervised}, where the authors proposed to treat each instance as a single class and extract features to distinguish different instances. Contrastive learning usually consists of three parts: anchor, positive samples, and negative samples. Anchor refers to a selected instance, positive samples denote samples that are similar to the anchor, and negative samples refer to samples that are dissimilar to the anchor. \cite{wu2018unsupervised} (see \textbf{Figure \ref{fig:moco}}) proposed a memory bank to store negative samples across batches which are essential to the generalization and diversity. In SwAV \cite{caron2020unsupervised}, the authors suggested comparing clustering centres, which has better semantic meaning than using a large number of negative samples as an approximation. MoCo \cite{he2020momentum, chen2020improved, chen2021empirical} and SimCLR \cite{chen2020simple, chen2020big} are the two foundational works that significantly advanced the boundaries of contrastive learning. MoCo proposed to utilize a momentum encoder in the network. The architecture has one encoder and one momentum encoder; the encoder is updated by gradient back propagation, while the momentum encoder is updated with the moving average from the encoder weights. In this siamese setting, the learning algorithm aims to force the representation from the encoder and the momentum encoder to be similar. In SimCLR, the authors stated that strong data augmentation is critical for contrastive learning. They also utilized a very large batch to have adequate negative samples during training. Most importantly, they proposed to use a non-linear projection head for the encoded feature and maximize the agreement for the projected representation. Additionally, they showed that using a larger model could achieve better results. in BYOL \cite{grill2020bootstrap}, the authors first proposed to perform contrastive learning without any negative samples. Normally, contrastive learning without negative samples will easily fall into trivial solutions as the network could easily force everything to be exactly the same as constant. BYOL borrowed core  principles from MoCo and SimCLR, and introduced a new predictor after the projection head. In SimSiam \cite{chen2021exploring}, the authors did not use a momentum encoder, negative samples, or large batches but still achieved comparable results to the previous studies. They proposed stop gradient operation, which is essential to the success of SimSiam to avoid mode collapse. In \cite{zbontar2021barlow}, the authors suggested measuring the cross-correlation between two identical networks and minimising the redundancy as the principle. In MoCo v3 \cite{chen2021empirical}, it used ViT as the backbone and further improved the performance. In DINO \cite{caron2021emerging}, it presented impressive results that the self-supervised learned attention is as good as the results of image segmentation. It is essentially an extension of BYOL with self-distillation \cite{hinton2015distilling}, forcing the learned representations from the teacher and the student networks to be similar.

\begin{figure*}[t]
  \centering
  \includegraphics[width=0.8\textwidth,height=9cm]{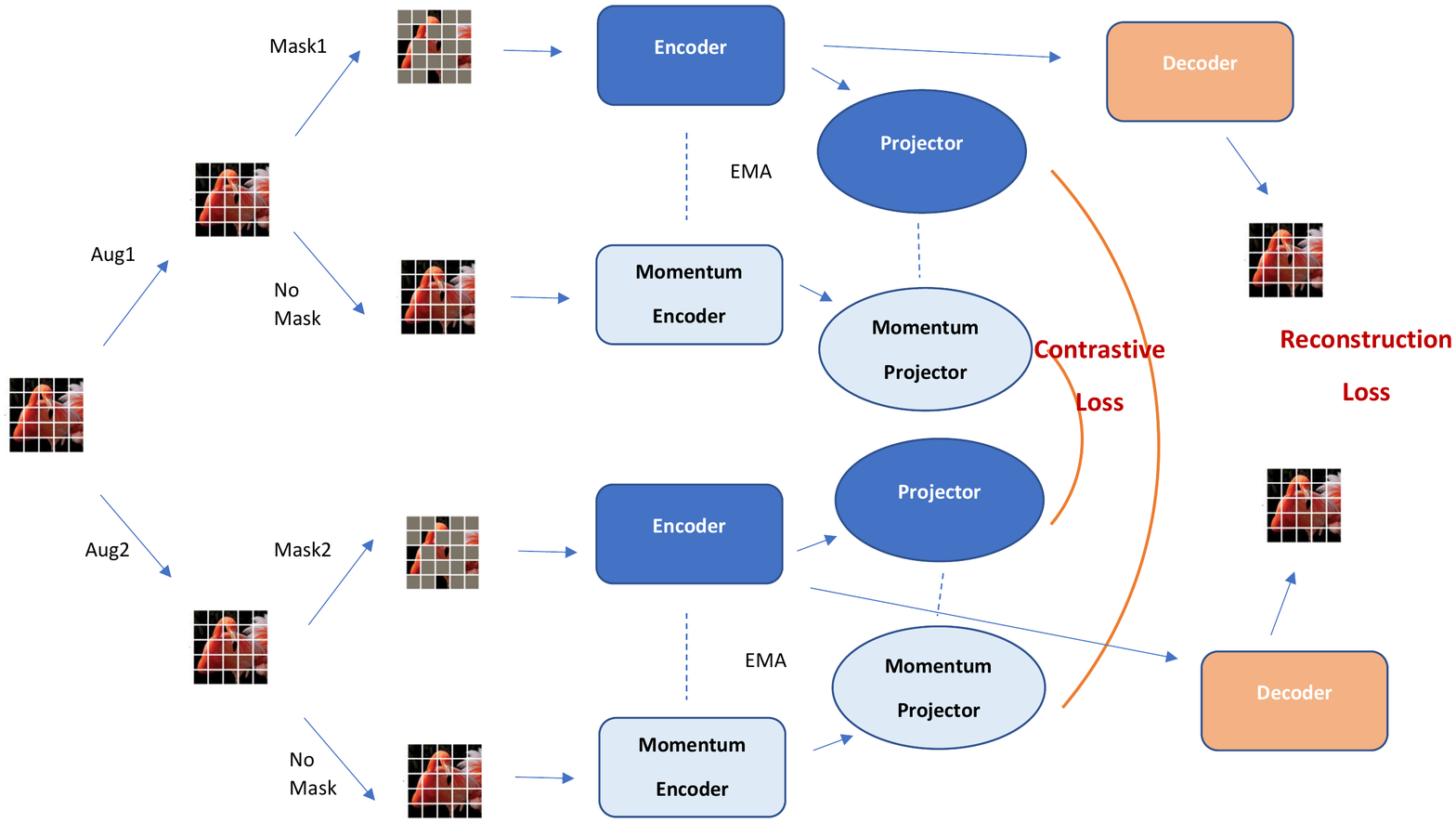}
   \caption{\textbf{The Overall Framework of MACRL.} The input image goes through two different data augmentation operations. Within each branch, there are two sub-branches, one applies high ratio of random masking, the other applies no mask. The same encoder and momentum encoder are used in those sub-branches, as well as the projector and momentum projector. The momentum components in the framework are all updated by exponential moving average, while the other components are updated by gradient back propagation. The same decoder is used to reconstruct the corrupted image. Overall, a contrastive loss (from the projected representations) and a reconstruction loss (from the decoded representations) are optimized in our proposed framework.}
   \label{fig:marcl_arch}
\end{figure*}

\section{Approach}
\label{sec:app}

Masked Contrastive Representation Learning (MACRL) is a self-supervised pre-training approach that builds upon masked image modelling and contrastive learning. The idea is straightforward: MACRL integrates masked image modelling into the contrastive learning framework, which is an asymmetric siamese network. The asymmetry refers to the difference in the strength of data augmentations and masking operations for the two branches of the siamese network. Overall, MACRL optimizes two objectives: a corruption reconstruction loss and a contrastive loss, for masked modelling and contrastive learning, respectively. The design of MACRL is illustrated in \textbf{Figure \ref{fig:marcl_arch}}.

\textbf{Asymmetric Siamese Network.} In MACRL, we adopt siamese encoder-decoder structure for the overall architecture, where there are two branches for each input sample: the main branch (contains the encoder and projector) and the momentum branch (contains the momentum encoder and momentum projector). Both branches share the same decoder for computing the masked image modelling loss since we only use the outputs from the encoder and projector to calculate the contrastive objective. We impose asymmetry in the siamese network setup, which follows the observations in \cite{wang2022importance}. Specifically, the source (i.e., main branch) should possess higher variance than the target (i.e., momentum branch). For the main branch, we apply extremely high mask ratio (e.g., 80\%) using a random masking strategy.
On the other hand, we reveal the original augmented image (i.e., 0\% mask ratio) for the momentum branch so that it possesses lower variance than the main branch. Furthermore, there are two sets of data augmentations, which follows the convention in BYOL \cite{grill2020bootstrap}. The difference in the augmentation strength introduces another level of asymmetric into the siamese network, which benefits the overall learning.

\textbf{Encoder.} The encoder is adapted from MAE \cite{he2022masked}, which is a vision transformer (ViT) \cite{dosovitskiy2020image} without prediction head (norm layer is included). As in \cite{chen2021empirical}, the authors pointed out that weight frozen in the patch embedding layer is essential for the stable training of ViT's contrastive learning, we also include the patch embedding frozen option in MACRL. Random masking is applied to the augmented samples, achieved by per-sample shuffling. The encoder generates the latent representation, the mask, and the index to restore the shuffling at the end. The learned representations in the encoder will be used as the final pre-trained weights. We did not freeze the patch embedding as suggested in \cite{chen2021empirical}. By default, the embedding dimension for the encoder is 512.

\textbf{Decoder.} We adopt asymmetric encoder and decoder setups, where the decoder is much shallower than the encoder. There is no need for a complex decoder as we will only use the learned representations (e.g., weights) from the encoder as the pre-training outcomes. Another reason for preferring the encoder over the decoder as pre-trained representation is that decoder is more likely to overfit to the pretext text (i.e., reconstruct the masked image). Whereas the encoder is more generalized. In MACRL, we utilize a 2-layer decoder, with one linear layer for matching the embedding space between the encoder and decoder and one attention layer for reconstructing the masked samples. Therefore, our decoder is even shallower and simpler than the one described in MAE. The decoder takes the latent representation and the index of shuffling from the encoder then outputs the reconstructed samples. By default, the embedding dimension for the decoder is 256

\textbf{Projector.} MACRL uses projector heads as described in other contrastive learning frameworks \cite{chen2021empirical, grill2020bootstrap}. The projector head is a two-layer feedforward network following \cite{chen2020simple}. We place Layer Normalization \cite{ba2016layer} instead of Batch Normalization \cite{ioffe2015batch} in the projector head as they work better with ViT encoder and our framework. This also results in a unified normalization layer in MACRL (i.e., all Layer-normalized). Unlike other contrastive learning approaches, MACRL does not place an additional predictor head (usually another shallow feedforward network) after the projector head for the encoder because we observed no performance gain. The latent representations from the encoder are first averaged in the token dimension and then passed to the projector head. For the momentum branch, the projector head is not updated by the gradient propagation (same as the momentum encoder); but by the exponential moving average. By default, the projected dimension is 512.

\textbf{Memory Bank.} We provide a memory bank as an option to store enough negative samples for contrastive learning objective when hardware resources constrain the batch size. According to the result in \cite{chen2021empirical}, MoCo achieves optimal results when the batch size is 4096, and there is no performance gain with increasing batch size. In our case, we notice that the memory bank works better in practice than the configurations with larger batch sizes (e.g., 4096) and without memory bank. By default, we set the memory bank size as 65,536.

\textbf{Corruption Reconstruction.} This is one of the two learning objectives in MACRL, which reconstruct the masked augmented images from the latent representations encoded by the encoder. The reconstruction is measured by $L_1$ distance ($L_2$ metric presents similar performance) between the original augmented sample and the reconstructed one. We only measure the distance for the visible tokens rather than the entire sample. The reconstruction helps MACRL to gain pixel-level representations from the data.

\begin{equation}
\mathcal{L}_{mim}=\mathcal{L}\left(\mathcal{D}_\theta \circ \mathcal{E}_\theta \left(\mathbf{x} \odot \mathrm{m}\right), \mathbf{x}\right)
\label{eq:mim}
\end{equation}

The objective is shown in \textbf{Equation \ref{eq:mim}}, there the $\mathcal{L}$ denotes distance measurement (e.g., $L_1$), $\mathcal{D}_\theta$ and $\mathcal{E}_\theta$ refer to the decoder and encoder, respectively. $\mathrm{m}$ is the mask applied to the input.

\textbf{Contrastive Objective.} MACRL adopts contrastive loss as the other learning objective. The contrastive criterion is based on InfoNCE \cite{oord2018representation}, which is shown in \textbf{Equation \ref{eq:cl}}.

\begin{equation}
\mathcal{L}_{cl}=-\log \frac{\exp \left(q \cdot k_{+} / \tau\right)}{\sum_{i=0}^K \exp \left(q \cdot k_i / \tau\right)}
\label{eq:cl}
\end{equation}

$q$ and $k$ are the normalized representations from the encoder and projector (and the momentum counterparts). A memory bank (i.e., $K$) is adopted to store adequate negative samples.  We apply the contrastive criterion on the normalized outputs from the projector heads and the momentum projector heads. A symmetric loss is imposed for the two different augmented views in both branches. The contrastive loss is scaled down by a factor of $\alpha$ (e.g., 10) to be within a similar range as the reconstruction loss. The contrastive objective enables MACRL to obtain high-level semantic information from the data. Both objectives are jointly optimized during training (see \textbf{Equation \ref{eq:loss}}).

\begin{equation}
\mathcal{L}_{macrl}= \alpha \times \mathcal{L}_{cl} + \mathcal{L}_{mim}
\label{eq:loss}
\end{equation}

\textbf{Simple Design.} MACRL is an end-to-end system with no separate stage, which is easy to implement. As shown in \textbf{Algorithm \ref{alg:macrl}}, the pseudo-code (PyTorch-like) for MACRL is concise and easily understandable. The base encoder is the main component in MACRL and is the final result (i.e., the pre-trained weights). We can elegantly plug in, play a vision transformer into the MACRL and obtain pre-trained attention layers.

\definecolor{codegreen}{rgb}{0,0.6,0}
\definecolor{codegray}{rgb}{0.5,0.5,0.5}
\definecolor{codepurple}{rgb}{0.58,0,0.82}
\definecolor{backcolour}{rgb}{0.95,0.95,0.92}

\lstdefinestyle{mystyle}{
    backgroundcolor=\color{backcolour},   
    commentstyle=\color{codegreen},
    keywordstyle=\color{magenta},
    numberstyle=\tiny\color{codegray},
    stringstyle=\color{codepurple},
    basicstyle=\ttfamily\footnotesize,
    breakatwhitespace=false,         
    breaklines=true,                 
    captionpos=b,                    
    keepspaces=true,                 
    numbers=left,                    
    numbersep=5pt,                  
    showspaces=false,                
    showstringspaces=false,
    showtabs=false,                  
    tabsize=2
}

\lstset{style=mystyle}
\begin{algorithm}
\caption{Masked Contrastive Representation Learning}\label{alg:macrl}
\begin{algorithmic}
\begin{lstlisting}[language=Python]
    
def MACRL(x):
    # e_q is the base encoder and p_q is the base projector
    # e_k is the momentum encoder and p_k momentum projector
    # m1 and m2 are two randomly generated masks
    # m0 is no mask
    # ctr is the contrastive objective
    # d is the decoder
    # rec is the reconstruction objective
    # alpha is the loss scaling factor
    
    x1, x2 = aug(x), aug(x)
    z1, z2 = e_q(x1, m1), e_q(x2, m2)
    q1, q2 = p_q(z1), p_q(z2)
    w1, w2 = e_k(x1, m0), e_k(x2, m0)
    k1, k2 = p_k(w1), p_k(w2)
    
    # contrastive loss
    cl_loss = ctr(q1 k2) + ctr(q2, k1)
    
    # decode
    x_1, x_2 = d(z1), d(z2)
    
    # reconstruction loss
    mim_loss = rec(x_1, x1) + rec(x_2, x2)
    
    return cl_loss * alpha + mim_loss
\end{lstlisting}
 \item
\end{algorithmic}
\end{algorithm}

\section{Experiments}
\label{sec:ex}

\textbf{Data Augmentation.} MACRL utilizes three different sets of data augmentation operations for pre-training, fine-tuning, and linear probing. For pre-training, each sample is mapped to two augmented views, with different augmentation strength. The operations follows BYOL \cite{grill2020bootstrap}, which has resize, crop, color jittering, greyscaling, Gaussian blurring, solarization, horizontal flip in the stronger augmented view, and the weaker augmented view has everything except solarization. As for fine-tuning, we apply random resize crop, AutoAugment \cite{cubuk2018autoaugment}, and CutOut \cite{devries2017improved}. For linear probing, we simply use random resize crop and random horizontal clip.

\textbf{Optimizer.} AdamW \cite{loshchilov2017decoupled} is the default optimizer for pre-training, fine-tuning, and linear probing. In the experiments, we also applied LARS \cite{you2017large} when the batch size is large (e.g., 4096), however, there is no significant performance gain and did not well during experiments. As for linear probing, we also used the Stochastic Gradient Descent (SGD) and LARS, but AdamW is still preferred. We use cosine annealing schedule \cite{loshchilov2016sgdr} for all experiments with warmup.

\textbf{Batch Size.} By default, we use memory bank in MACRL. Therefore, there is no need for large batch size to account for adequate negative examples that are essential for contrastive objective. Hence, we use batch size of 2,048 by default. Due to the limitation of resource, we could not set the batch size to 2,048 for large dataset directly. In that case, we adopt accumulate gradient to mimic large batch size.

\begin{table}
  \centering
  \begin{tabular}{@{}p{2cm}lc@{}p{1cm}lc@{}}
    \toprule
    Method & CIFAR-10 & CIFAR-100 \\
    \midrule
    MAE \cite{he2022masked} & 96.18 & 81.68\\
    MoCo \cite{he2020momentum} & 95.61 & 74.59\\
    MACRL$^{\ast}$ & 96.43 & 81.38 & \\
    MACRL & \textbf{97.88}  & \textbf{82.94}\\
    \bottomrule
  \end{tabular}
  \caption{\textbf{Fine-tuning Results on CIFAR}. The fine-tuning on CIFAR-10 and CIFAR-100 with MAE, MoCo, and MACRL. The $\ast$ denotes the results for fine-tuning 100 epochs.}
  \label{tab:cifar-ft}
\end{table}

\begin{table}
  \centering
  \begin{tabular}{@{}p{2cm}lc@{}p{1cm}lc@{}}
    \toprule
    Method & CIFAR-10 & CIFAR-100 \\
    \midrule
    MAE \cite{he2022masked} & 78.60 &50.08\\
    MoCo \cite{he2020momentum} & 86.16 & 57.71\\
    MACRL$^{\ast}$ & 81.36 & 48.62 & \\
    MACRL & \textbf{91.02} & \textbf{66.27}\\
    \bottomrule
  \end{tabular}
  \caption{\textbf{Linear Probe Results on CIFAR}. The linear probing on CIFAR-10 and CIFAR-100 with MAE, MoCo, and MACRL. The $\ast$ denotes the results for linear probed 100 epochs.}
  \label{tab:cifar-lp}
\end{table}

\textbf{Pre-Training.} For pre-training, only the images are used for training the model, and the labels are not accessible. At the end of the pre-training, we extract the encoder (i.e., the pre-trained visual transformer) from the MACRL's main branch and store it for fine-tuning and linear probing evaluations. By default, we pre-trained the model for 2,000 epochs with 50 warm-up epochs. The learning rate is $1.5\mathrm{e}{-4}$ with 0.01 weight decay. The mask ratio is, by default set to 0.75.

\textbf{Fine-Tuning.} In fine-tuning, we load the pre-trained weight to a visual transformer and train the entire model end-to-end with both images and the corresponding labels. By default, the learning rate is $1.5\mathrm{e}{-3}$ with 0.01 weight decay.

\textbf{Linear Probing.} Similar to fine-tuning, we load the pre-trained weight to a visual transformer in linear probing. However, all the weights are frozen except for the prediction head. We only train the prediction head with labelled images. We set the learning rate as 0.1. The weight decay is set to zero, following the setups in MAE \cite{he2022masked}.

\subsection{CIFAR}

We first evaluated MACRL on two standard benchmarks: CIFAR-10 and CIFAR-100 \cite{krizhevsky2009learning}, which both contain 60,000 small-size images, respectively. CIFAR-10 has 10 classes, and CIFAR-100 has 100 classes that are more difficult. We used 12-Layer (4 heads) encoder and a single attention layer (1 head) decoder in MAE, MoCo and MACRL and CIFAR datasets. Patch size is set to 4 for both datasets. The fine-tuning and linear probe results are shown in \textbf{Table \ref{tab:cifar-ft}} and \textbf{Table \ref{tab:cifar-lp}}, respectively.

To get the results shown in the tables, all three methods (i.e., MAE, MoCo, MACRL) were pre-trained for 2,000 epochs for both datasets. For MAE and MoCo, we fine-tuned and linear probed for 400 and 1,000 epochs, respectively. As for MACRL, we fine-tuned for 200 epochs and linear probed for only 200 epochs. Furthermore, MACRL could already achieve comparable or better results when only fine-tuned or linear probed for 100 epochs. Therefore, results prove that MACRL has better performance than MAE and MoCo. More importantly, the learned representations from MACRL can be tuned to higher accuracy easier and much more efficiently than the other two methods.

\begin{figure*}
  \centering
  \begin{subfigure}{0.48\linewidth}
    \centering
    \includegraphics[width=0.8\linewidth]{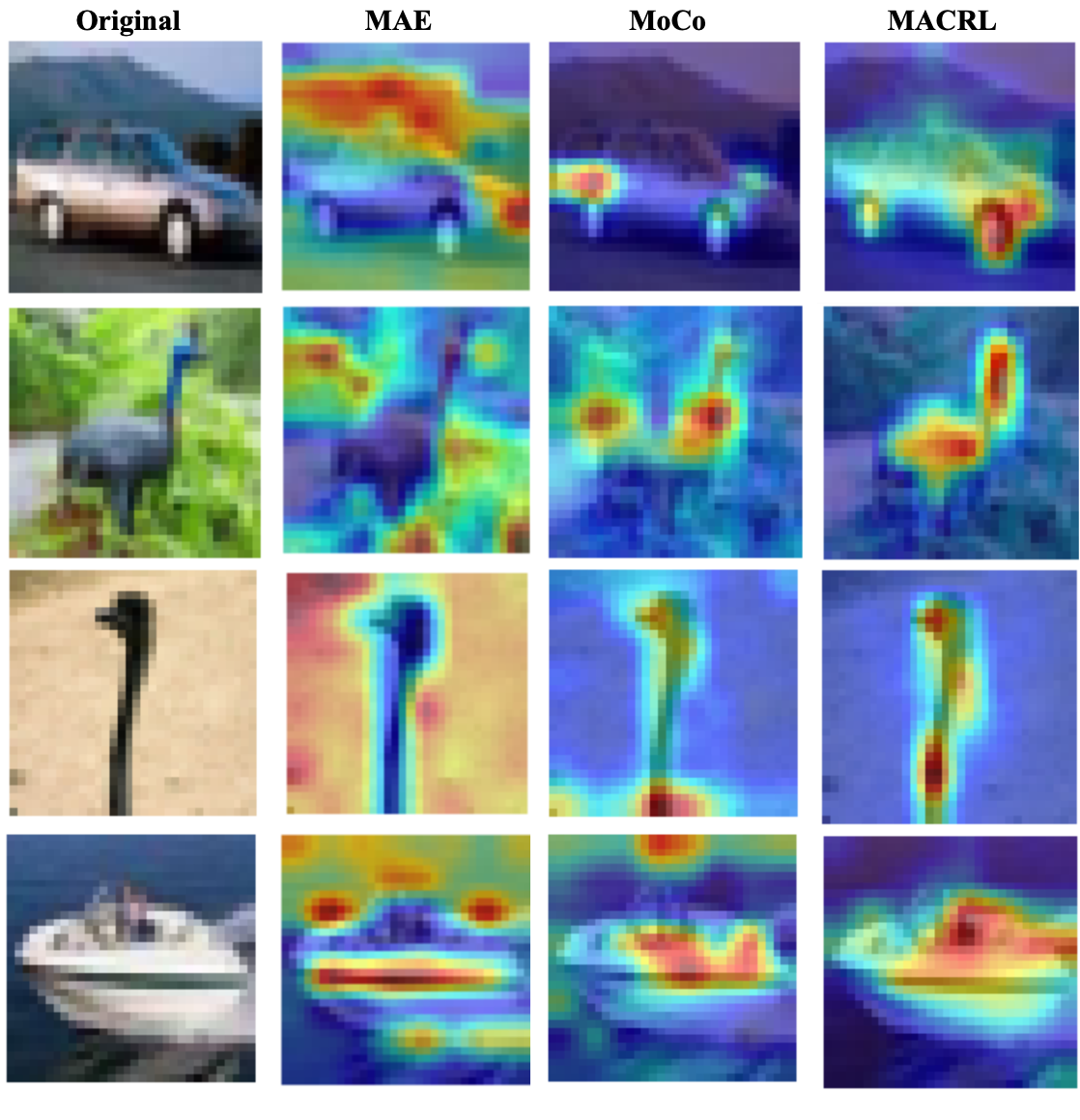}
    \caption{Attention Visualization on CIFAR Dataset}
    
    \label{fig:att_cifar}
  \end{subfigure}
  \hfill
  \begin{subfigure}{0.49\linewidth}
  \centering
    \includegraphics[width=0.81\linewidth]{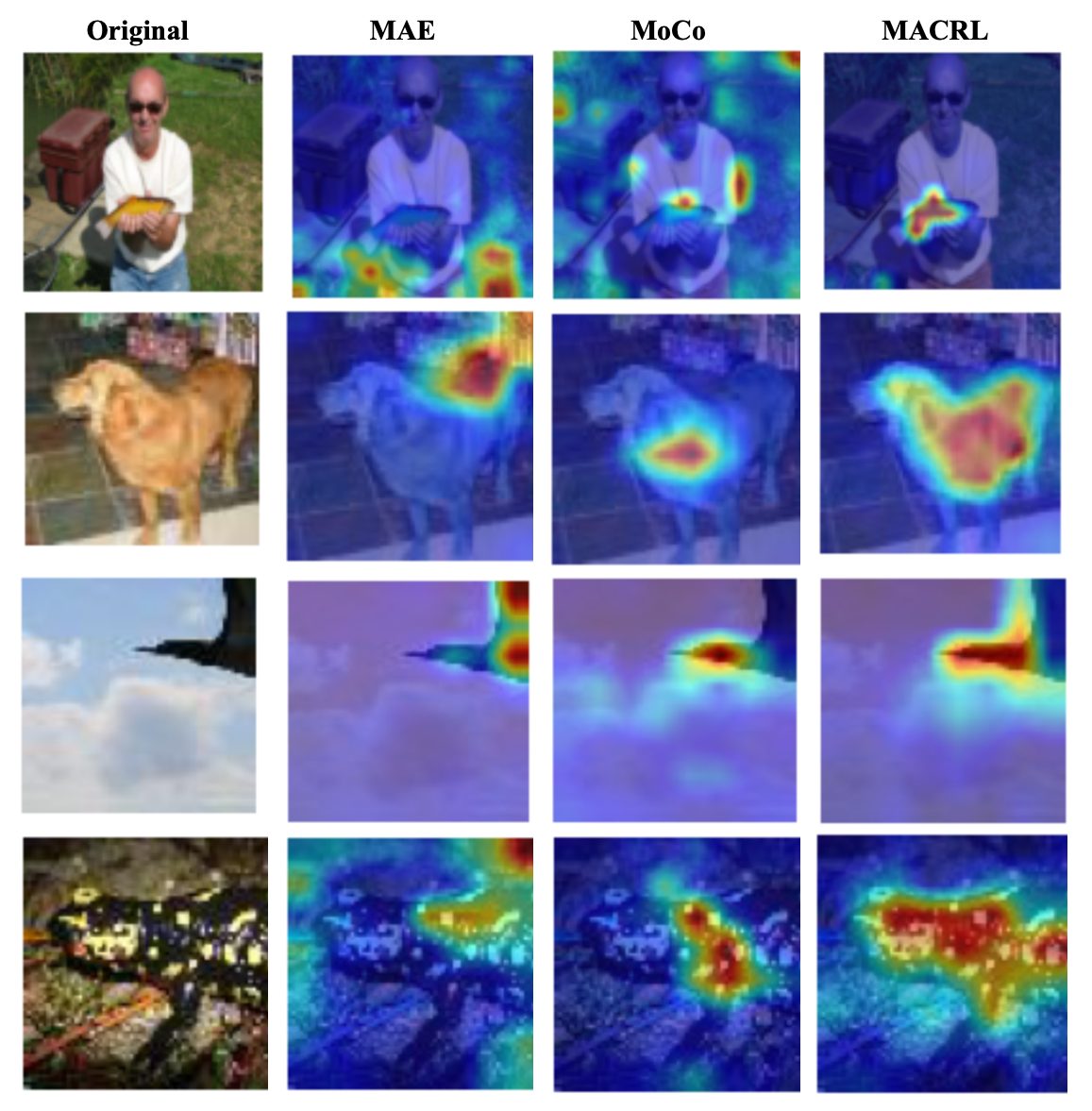}
    \caption{Attention Visualization on ImageNet Subsets.}
    \label{fig:att_img}
  \end{subfigure}
  \caption{\textbf{Attention Visualization}. We applied the method described in \cite{chefer2021transformer} to interpret the learned representations from three self-supervised learning approaches. \textbf{Figure \ref{fig:att_cifar}} shows the results on CIFAR-10 and CIFAR-100. \textbf{Figure \ref{fig:att_img}} presents the results on Imagenette (see row 1) and Tiny-ImageNet (see row 2, 3, 4).}
  \label{fig:att}
\end{figure*}

\begin{table}
  \centering
  \begin{tabular}{@{}lc@{}lc@{}lc@{}}
    \toprule
    Method & Tiny-ImageNet { }{ } & Imagenette & Imagewoof \\
    \midrule
    MAE \cite{he2022masked} & 70.56 & 92.86 & 84.47\\
    MoCo \cite{he2020momentum} & 70.08 & 92.26 & 83.28\\
    MACRL$^{\ast}$ & 69.86& 92.89 &  81.21 \\
    MACRL & \textbf{72.06} & \textbf{93.37} & \textbf{87.14}\\
    \bottomrule
  \end{tabular}
  \caption{\textbf{Fine-tuning Results on ImageNet Subsets}. The fine-tuning on ImageNet subsets with MAE, MoCo, and MACRL. The $\ast$ denotes the results for fine-tuning 200 epochs.}
  \label{tab:img-ft}
\end{table}

\begin{table}
  \centering
  \begin{tabular}{@{}lc@{}lc@{}plc@{}}
    \toprule
    Method & Tiny-ImageNet { }{ } & Imagenette & Imagewoof \\
    \midrule
    MAE \cite{he2022masked} & 61.95 & 74.26 & 44.74\\
    MoCo \cite{he2020momentum} & 75.25 & 78.60 & 51.53\\
    MACRL$^{\ast}$ & 54.45& 71.05 &  44.76 \\
    MACRL & \textbf{75.44} & \textbf{80.56} & \textbf{54.98}\\
    \bottomrule
  \end{tabular}
  \caption{\textbf{Linear Probe Results on ImageNet Subsets}. The fine-tuning on ImageNet subsets with MAE, MoCo, and MACRL. The $\ast$ denotes the results for linear probed 500 epochs.}
  \label{tab:img-lp}
\end{table}

\subsection{ImageNet Subsets}

ImageNet-1K (IN1K) \cite{deng2009imagenet} is the de-facto dataset for benchmarking visual models. We trained the model on several subsets from IN1K, including Tiny ImageNet \cite{le2015tiny}, Imagenette and Imagewoof \cite{howard2020fastai}. Tiny ImageNet contains $64 \times 64$ images from 200 classes, while Imagenette and Imagewoof contain 10 easy classes and 10 difficult classes from the original ImageNet-1K, respectively. We use 8 as the patch size for Tiny-ImageNet and 16 as the patch size for Imagenette and Imagewoof. For Tint ImageNet, we used the same network configurations as CIFAR-10 and CIFAR-100. Since Imagenette and Imagewoof have smaller scales, we adopted a 4-layer encoder (4 heads) and a single attention layer decoder (1 head) for the experiments for all three approaches. The results are shown in \textbf{Table \ref{tab:img-ft}} and \textbf{Table \ref{tab:img-lp}}, respectively.

We pre-trained 2,000 epochs for methods on each dataset. For all three methods, we fine-tuned for 400 epochs, and linear probed for 1,000 epochs. Similar to the findings in CIFAR-10 and CIFAR-100, MACRL can achieve comparable or even better results than MAE and MoCo when only fine-tuned for 200 epochs. As for Linear Probe, MACRL also presents superior performance when tuned with the same epochs as the other two methods and can achieve comparable results when tuned only with half the total epochs. Therefore, the results on the ImageNet subsets align with our previous observations and validate the performance and efficiency of MACRL over the other two approaches.

\section{Discussion}
\label{sec:discuss}

According to the experiment results, MACRL shows better performance across different benchmarks. Moreover, we show that the learned representations from MACRL can be tuned more easily and efficiently (e.g., better accuracy within less epochs). Furthermore, we visualize the attention from the pre-trained weights using the method described in \cite{chefer2021transformer}. According to the results shown in \textbf{Figure \ref{fig:att_cifar}} and \textbf{Figure \ref{fig:att_img}}, we can see that MACRL has better interpretability than MAE and MoCo as MACRL focuses on the objects in the image, especially the key components. However, MAE and MoCo present scattered attention over the image and do not have a clear emphasis on the salient objects. Furthermore, the visualizations show that MAE and MoCo focus more on the background in the image rather than the foreground object. Instead, MACRL pays more attention to only the foreground objects. Additionally, we observed performance drops (e.g.,fine-tune, linear probe accuracy, and visualizations) for MAE and MoCo when they are applied on smaller scale datasets using lightweight network structure. Whereas, MACRL still maintains good performance given smaller-scale data and smaller model size.

We believe that the power of MACRL comes from combining the merits of contrastive learning and masked modelling. As shown by \cite{xie2022revealing}, the self-supervised learned feature from contrastive learning has very high linear probe accuracy, which is very similar to supervised learning. Moreover, \cite{chen2021empirical} showed that using a plain k-Nearest Neighbour (kNN) classifier after the final layer of contrastive learning pre-trained ViT could achieve decent performance. Whereas, the kNN accuracy for MAE is poor. On the other hand, the attention from masked modelling are more diverse (i.e., attention heads), where each layer has a similar attention distance. On the contrary, the attention head becomes less diverse as the layer goes deeper for contrastive learning. This indicates that masked modelling is better at aggregating different kinds of information (e.g., local vs global, focus vs broad). Therefore, by integrating both contrastive learning and masked modelling, MACRL gains the representation which mimics the supervised learning at the last layer and is also equipped with diverse representations over different intermediate layers. 

\section{Conclusion}
\label{sec:con}

In this work, we present a new self-supervised representation method that combines two mainstream approaches: masked modelling and contrastive learning. We show that integrating those two mainstream approaches yields more powerful representations than when are used individually. The proposed MACRL framework shows superior performance against five different benchmarks than the existing methods in both fine-tuning and linear probe. Furthermore, the results validate the efficiency and interpretability of our proposed approach, where MACRL can achieve better accuracy with less epochs and generate more semantically meaningful attention. The main purpose of this work is to introduce a novel approach for self-supervised representation learning by unifying the existing two mainstream methods, rather than benchmarking with the state-of-the-art.  In the future, we plan to justify the performance of MACRL on a larger scale datasets (e.g., complete IN1K) and investigate the transferability on downstream tasks (e.g., detection and segmentation).

\section{Acknowledgements}
The authors of this paper would like to thank the hardware support for all the experiments. This research was undertaken using the LIEF HPC-GPGPU Facility hosted at the University of Melbourne. This Facility was established with the assistance of LIEF Grant LE170100200.
{\small
\bibliographystyle{ieee_fullname}
\bibliography{egbib}
}

\end{document}